\documentclass[conference, 10 pt]{IEEEtran}
\usepackage{cite}
\usepackage{amsmath,amssymb,amsfonts}
\usepackage{algorithmic}
\usepackage[ruled]{algorithm2e}
\usepackage{graphicx}
\usepackage{textcomp}
\usepackage{comment}
\usepackage{xcolor}
\usepackage{multirow}
\usepackage{caption}
\usepackage{booktabs}
\usepackage{colortbl}
\usepackage{makecell}
\usepackage{subcaption}
\usepackage{enumitem}
\usepackage{xcolor}
\usepackage{hyperref}

\begin{document}

\title{STRAP-ViT: \underline{S}egregated  \underline{T}okens with  \underline{R}andomized - Transformations for Defense against  \underline{A}dversarial  \underline{P}atches in \underline{ViT}s}

\author{Nandish Chattopadhyay $^{1}$, Anadi Goyal $^{1}$, Chandan Karfa $^{1}$ and Anupam Chattopadhyay $^{2}$ \\
$^1$ Indian Institute of Technology, Guwahati, India and 
$^2$ Nanyang Technological University, Singapore}

\maketitle

\begingroup
\renewcommand\thefootnote{\fnsymbol{footnote}}
\footnotetext{This paper has been accepted for publication at the 63rd IEEE/ACM Design Automation Conference (DAC) 2026 in Long Beach, California, USA. }
\endgroup

\begin{abstract}
Adversarial patches are physically realizable localized noise, which are able to hijack Vision Transformers (ViT) self-attention, pulling focus toward a small, high-contrast region and corrupting the class token to force confident misclassifications. In this paper, we claim that the tokens which correspond to the areas of the image that contain the adversarial noise, have different statistical properties when compared to the tokens which do not overlap with the adversarial perturbations. We use this insight to propose a mechanism, called STRAP-ViT, which uses Jensen-Shannon Divergence as a metric for segregating tokens that behave as anomalies in the Detection Phase, and then apply randomized composite transformations on them during the Mitigation Phase to make the adversarial noise ineffective. The minimum number of tokens to transform is a hyper-parameter for the defense mechanism and is chosen such that at least $50\%$ of the patch is covered by the transformed tokens. STRAP-ViT fits as a non-trainable plug-and-play block within the ViT architectures, for inference purposes only, with a minimal computational cost and does not require any additional training cost/effort. STRAP-ViT has been tested on multiple pre-trained vision transformer architectures (ViT-base-16 \cite{VIT} and DinoV2 \cite{dinov2}) and datasets (ImageNet \cite{imagenet} and CalTech-101 \cite{caltech}), across multiple adversarial attacks (Adversarial Patch \cite{googleap}, LAVAN \cite{lavan}, GDPA \cite{gdpa} and RP2 \cite{RP2}), and found to provide excellent robust accuracies lying within a $2-3\%$ range of the clean baselines, and outperform the state-of-the-art. 

\end{abstract}

\section{Introduction}
\label{introduction}
Vision Transformers (ViTs) are central to next-gen AI because the same token-based architecture that scaled language now scales vision—and increasingly, multi-modal learning—under a unified recipe for pretraining, transfer, and deployment \cite{VIT, dinov2}. Their global self-attention captures long-range structure essential for video, 3D perception, medical imaging, remote sensing, and robotics, while hierarchical/windowed variants and sparsity make them hardware-friendly from data centers to edge/FPGA targets. Building and deploying such models demands enormous capital: recent analyses show training costs for frontier-scale models growing around 2.4x per year with single runs trending toward hundreds of millions, while leading labs reportedly spent several billions on compute in 2024 alone. Top companies are pouring tens of billions of dollars per quarter into AI infrastructure (chips, servers, and data centers) to build and run large multimodal/vision models—e.g., Microsoft spent ~USD 35B in Q3 2025, Amazon USD 34.2B, Alphabet guided USD 91–93B for full-year 2025 capex, and Meta plans USD 66–72B in 2025 \cite{BI, investopedia}. Given these stakes, rigorous security research is critical: ViTs remain vulnerable to physically realizable adversarial patches that can hijack attention and cause reliable mispredictions, so hardening them (training, detection, certification) is essential to protect real-world deployments and the capital invested in them. 

Specifically, adversarial attacks—especially localized adversarial patches—pose a severe, practical risk to Vision Transformers (ViTs) because a small, printable pattern can hijack self-attention and dominate the class token, causing confident misclassification even when the patch covers only a tiny fraction of the image \cite{Goodfellow2015ExplainingAH}. Unlike CNNs, ViTs’ weak built-in locality and global token mixing make them particularly susceptible to attention redirection: a high-contrast patch can soak up attention across layers, corrupt token interactions, and transfer across models and tasks. These patches are robust to common preprocessing, survive viewpoint and lighting changes, and work in the physical world (stickers, clothing, signage) \cite{Hu21, googleap, lavan, gdpa}. Even windowed/hierarchical variants (e.g., Swin transformers \cite{swin}) remain vulnerable through boundary and cross-window effects. While defenses (adversarial training, patch detectors/maskers, token-level smoothing, certified radii) help, many are brittle or compute-heavy—leaving ViTs exposed in safety-critical deployments unless patch-aware robustness is treated as a first-class objective.

\begin{figure}[!htbp]
\centerline{\includegraphics[width=\columnwidth]{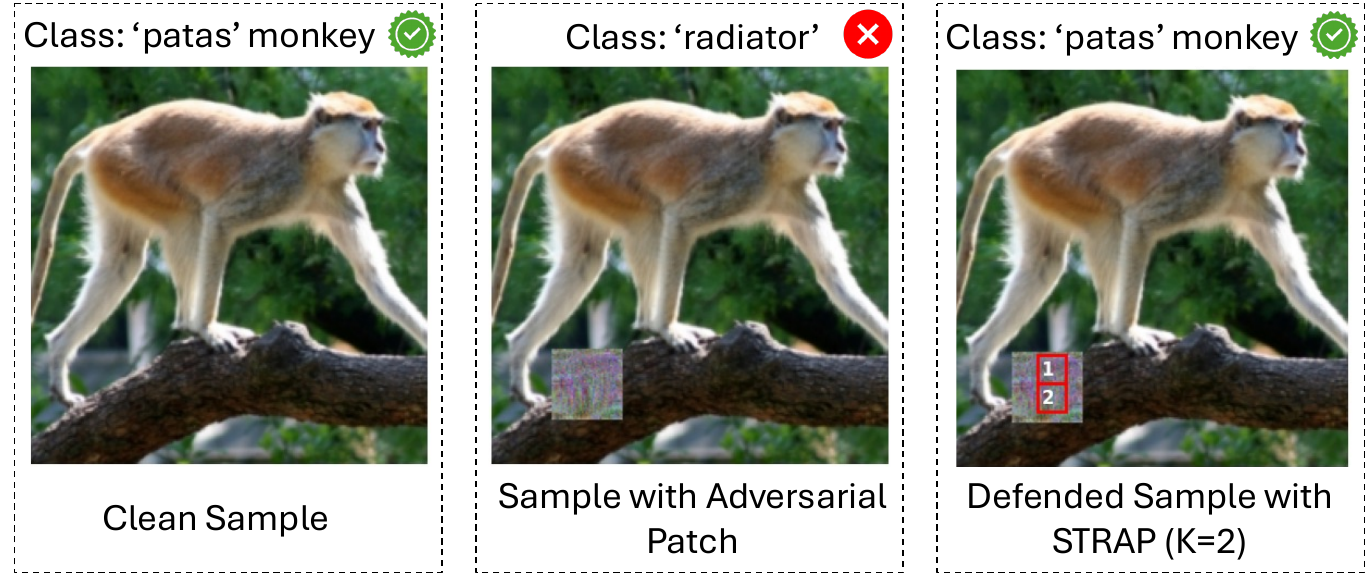}} 
\caption{STRAP-ViT in action: Demonstrating the detection and transformation of tokens which overlap with the adversarial patch using the proposed STRAP-ViT defense technique. In this case, the minimum number of tokens to modify is two ($K=2$), such that at least $50\%$ of the patch is covered, to render the patch ineffective. }
\label{fig:demo}
\end{figure}

\subsection{Motivation}
\begin{figure}[htbp]
\centerline{\includegraphics[width=\columnwidth]{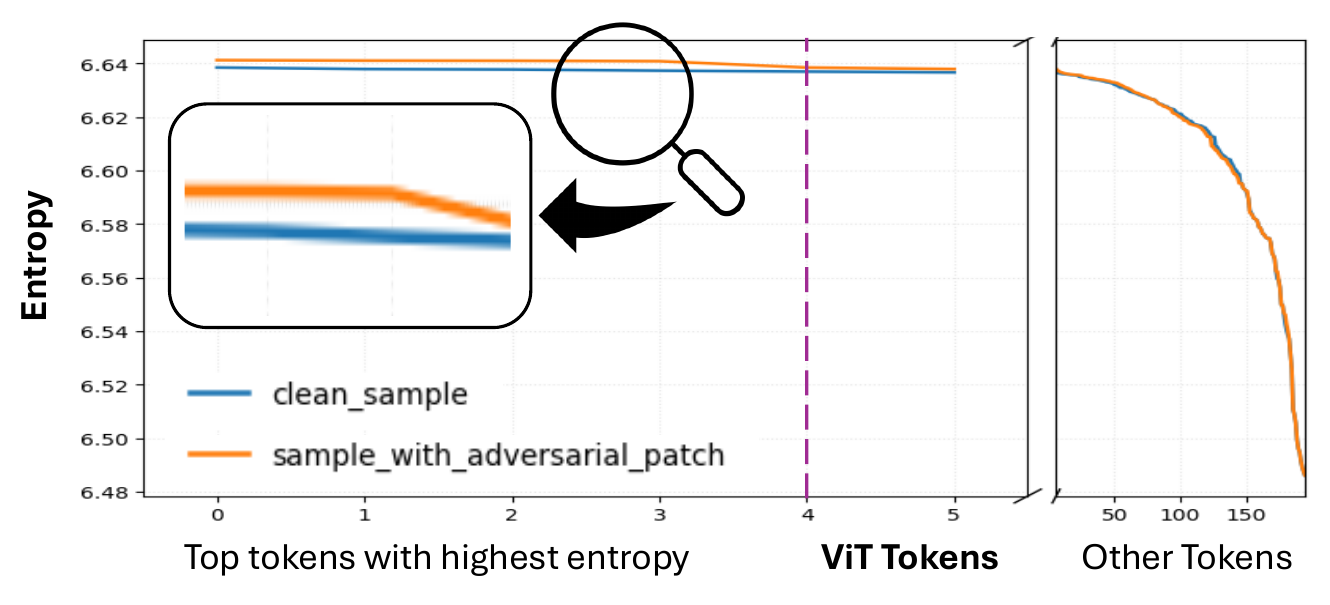}} 
\caption{ Illustration of the statistical separability of the tokens which overlap with the adversarial patch and other tokens from the clean part of the image, using entropy as a metric.  }
\label{fig:entropy}
\end{figure}

Adversarial patches perturb only a subset of vision tokens, yet those tokens exhibit a distinct informational signature: when a patch overlaps a token’s receptive field, the token’s channel distribution systematically shifts and its Shannon entropy \cite{entropy} rises relative to non-overlapping tokens, as shown in Figure \ref{fig:entropy}. This empirical regularity motivates a localization strategy that treats entropy as a preliminary indicator and uses stronger, symmetric discrepancy measures—specifically the Jensen–Shannon divergence (JSD) between a token’s induced probability map and a clean/reference distribution—to identify patch-affected regions \cite{JSD1}. Since JSD, particularly $(\sqrt{\mathrm{JSD}})$ is a true metric and tightly linked to mutual information, elevated scores provide calibrated evidence of patch influence rather than generic uncertainty. Building on these scores, the method selectively neutralizes high-JSD tokens via randomized composite transformations (e.g., $(L_p)$ projections, bounded affine contractions, and temperature-scaled softmax \cite{freedman}), while leaving unaffected tokens unchanged. It evaluates anomalies in the same semantic space the model uses for prediction, making detection more aligned with the model’s internal reasoning. The result is a targeted, model-agnostic defense that improves ViT robustness with minimal disruption to clean content, shown in Figure \ref{fig:demo}.

\vspace{-3mm}

\subsection{Contributions}
The contributions of this paper are:
\begin{itemize}
    \item \textbf{STRAP-ViT} introduces a novel approach of making ViTs robust against adversarial patches, by modifying selected tokens within the ViT inference flow  (without any additional training cost). This makes use of the statistical properties of the informational content present in the adversarial patch, and its difference to the distribution of information in the clean image, to successfully identify and eliminate the adversarial behaviour. 
    \item \textbf{STRAP-ViT} is a model and patch agnostic algorithm which uses statistical metrics for identifying the location of the adversarial patch using the Detection through Token Segregation process, which makes use of Jensen-Shannon Divergence. Thereafter, the Mitigation through Token Transformation process uses a random composition of transformations to eliminate adversarial noise, as presented in Section \ref{methodology}.  
    \item \textbf{STRAP-ViT} has been thoroughly evaluated with multiple pre-trained vision transformer architectures (ViT-base-16 and DinoV2 backbone model) and datasets (ImageNet and CalTech-101), across multiple adversarial attacks (Adversarial Patch \cite{googleap}, LAVAN \cite{lavan}, GDPA \cite{gdpa} and RP2 \cite{RP2}). STRAP-ViT is able to provide robustness by generating accuracies lying within a $2-3\%$ range of the baselines across different settings, without a drop in the clean case accuracies, and outperforms the state-of-the-art, as presented in Section \ref{results}. 
\end{itemize}

\section{Background and Related Work}
\label{background}
In this section, we describe the class of deep neural networks which are considered in this paper, the mechanism of generating the adversarial attack in the form of adversarial patches, the overall threat model in which the attacker operates to set the context for the proposed adversarial defense technique and the related works in adversarial defenses.  

\subsection{Vision Transformers}
Vision Transformers (ViTs) treat an image as a sequence: they split it into non-overlapping patches (e.g., 16×16), flatten each patch, linearly project to a D-dim embedding, add a learnable class token and positional encodings, then feed the sequence to a stack of Transformer encoder blocks. Each block uses pre-norm LayerNorm, multi-head self-attention (MSA), and a position-wise MLP, with residual connections around both MSA and MLP. The class token (or a global average over tokens) provides the image representation for classification.  

\subsection{Adversarial Patches}
Adversarial patches are localized, intentionally crafted regions added to an image to cause deep learning models—particularly vision architectures like CNNs and Vision Transformers (ViTs)—to misclassify inputs with high confidence, even when the rest of the image remains unaltered. Unlike pixel-level perturbations that require full-image access, patch attacks are physically realizable: a small printed sticker or pattern can reliably fool models under various viewing conditions. These patches often exploit the model’s overreliance on salient spatial features,hijacking attention in ViTs or convolutional feature maps, redirecting focus away from true object regions. Because they are input-agnostic and robust across contexts, adversarial patches pose a severe threat to real-world applications such as surveillance, autonomous driving, and facial recognition \cite{googleap, lavan}. Formally, an adversarial patch for ViTs can be created in the method described as follows.
An adversarial patch $(p\in[0,1]^{k\times k\times 3})$ is optimized to force a model $(f)$ to predict a target $(y_t)$ when pasted onto images $(x\sim\mathcal D)$. Let a placement operator $(S(x,p,\tau)= (1-M_\tau)\odot x + M_\tau\odot \Pi(p))$, where $(M_\tau)$ is a binary mask after transform $(\tau\in\mathcal T)$ (scale/rotate/translate) and $(\Pi)$ resizes $(p)$. The Expectation Over Transformation (EOT) objective is
\[
p^\star=\arg\max_{p\in[0,1]^{k\times k\times 3}}\ \mathbb E_{x\sim\mathcal D,\ \tau\sim\mathcal T}\ \ell\big(f(S(x,p,\tau)),,y_t\big),
\]
with projected gradient ascent step
\[
p_{t+1}=\Pi_{[0,1]}\Big(p_t+\alpha,\operatorname{sign}\big(\nabla_p\ \mathbb E_{x,\tau}\ \ell(f(S(x,p,\tau)),y_t)\big)\Big).
\]
For ViTs, the patch induces attention hijack by inflating attention toward its token $(m)$: with $(A=\operatorname{softmax}\big(QK^\top/\sqrt{d}\big))$, the attack increases $(A_{\cdot m})$, effectively steering the class token’s representation and yielding high-confidence misclassification, making this form of attack a very potent one.

\begin{figure*}[htbp]
\centerline{\includegraphics[width=2\columnwidth]{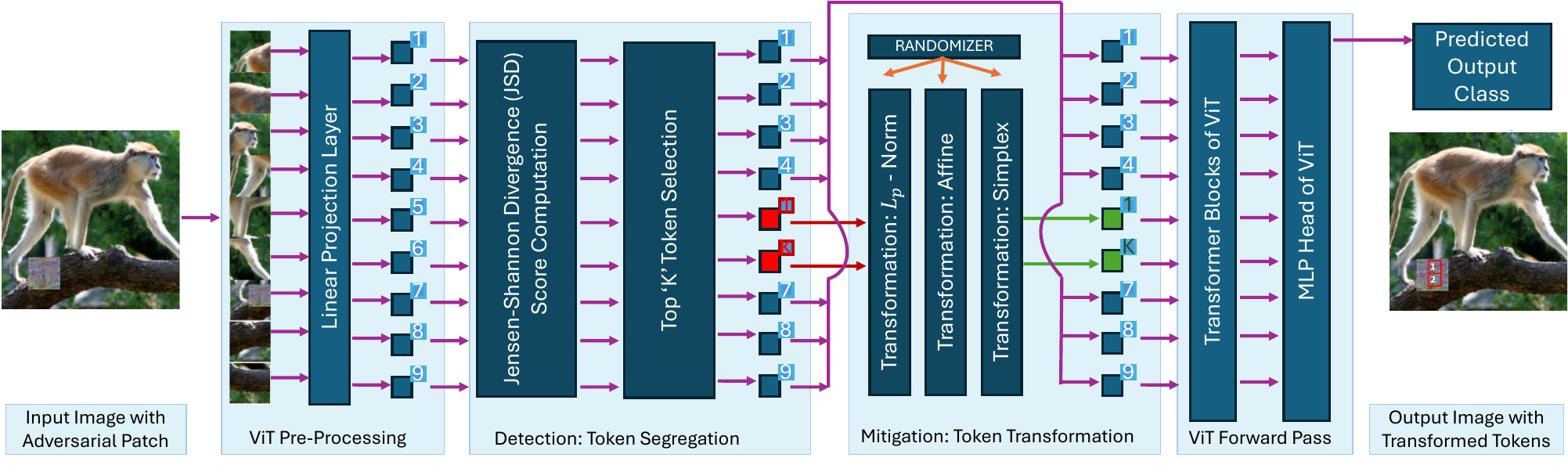}} 
\caption{STRAP-ViT pipeline: The two-stage adversarial defense framework comprising of the Detection using Token Segregation through Jensen-Shannon Divergence and Mitigation using Token Transformation through composite randomised transformations, preceded by ViT Pre-processing Tokenization and succeeded by Vit Forward Pass.   }
\label{fig:pipeline}
\end{figure*} 

\subsection{Threat Model}
We operate under the strongest assumption, a white-box scenario, where the attacker possesses complete knowledge of the victim DNN, including its architecture and parameters. As in other proposed defenses \cite{levine2020randomized, xiang2021patchguard, naseer2019local}, the adversary is capable of substituting a specific region of an image with an adversarial patch, with the goal of consistently inducing misclassification / misdetection / misestimaion across all input instances. Specifically, the threat of a white-box adversarial attack with a pre-trained adversarial patch has the following considerations:
\begin{itemize}[leftmargin=*]
    \item Localization: Only a bounded region (e.g., $\leq 5\%$ of the image) is modified.
    \item Transferability: The same patch generalizes across different images and across models.
    \item Physical realizability: The patch can be printed and photographed, maintaining its adversarial effect under real-world lighting, perspective, and noise.
\end{itemize}

\subsection{Literature Review: Adversarial Defenses}
Defences against adversarial attacks can be broadly divided into certifiable \cite{chiang2020certified, chen2022towards, xiang2022patchcleanser, salman2022certified} and empirical approaches \cite{gittings2020vax, rao2020adversarial, hayes2018visible, naseer2019local, xiang2021patchguard, cod_1, cod_2, cod3, features}. Certifiable methods provide formal guarantees that predictions remain stable under any valid patch placement. Salman et al. \cite{salman2022certified} show that smoothed Vision Transformers can substantially strengthen certified patch robustness while preserving near-standard accuracy and offering much faster inference. Chen et al. \cite{chen2022towards} adapt Derandomized Smoothing to ViTs using progressive smoothed modeling and structured attention, achieving practical robustness with limited clean-accuracy loss. Xiang et al. propose PatchCleanser \cite{xiang2022patchcleanser}, a classifier-agnostic defence that isolates and suppresses potential patch regions via two-stage masking and provides certified guarantees for any downstream model. Despite these advances, certifiable approaches remain computationally demanding and often require specialised architectures or training, constraining their practicality.
In contrast, empirical defences do not provide formal guarantees but are generally more flexible and easier to deploy. One popular line of work is adversarial training \cite{gittings2020vax, rao2020adversarial}, where models are trained against patch perturbations, enabling models to learn patch-resilient features. However, such training is computationally heavy and must be repeated for different threat models. Another direction focuses on patch detection and suppression \cite{jedi, chattopadhyay2025oddr, ChattopadhyayGH24, chattopadhyay2024anomaly, naseer2019local, xiang2021patchguard}, motivated by the localised high-frequency artifacts introduced by patches. Jedi \cite{jedi} identifies adversarial patches by exploiting their characteristically high local entropy and uses an autoencoder to complete high-entropy regions, enabling precise localization and restoration. DefensiveDR \cite{ChattopadhyayGH24} mitigates patch effects by projecting images into a lower-dimensional space, suppressing localized perturbations while retaining essential task-relevant structure. ODDR \cite{chattopadhyay2025oddr} segments an image, flags segments whose features deviate statistically from the natural distribution, and then neutralizes these outlier regions through reduced-representation smoothing.  
Unlike these prior defenses that operate purely in pixel space, our proposed approach leverages the ViT’s native token representations, allowing adversarial perturbations to be detected directly in the semantic feature space where the model performs reasoning. Operating at the token level makes our proposed defense inherently more suitable for ViT architectures.



\section{STRAP-ViT Defense Mechanism}
\label{methodology}

In this Section, we describe the entire algorithm of the adversarial defense and explain each component, which include the Detection using Token Segregation through Jensen-Shannon Divergence and Mitigation using Token Transformation through composite randomized transformations, and how they fit into the inference pipeline of the ViT architecture, as shown in Figure \ref{fig:pipeline}. 
\begin{itemize}[leftmargin=*]
    \item After the patch embedding layer in the ViT architecture converts image patches into tokens, the token segregation algorithm calculates the Jensen-Shannon Divergence \cite{JSD1}, specifically the $(\sqrt{\mathrm{JSD}})$ for each token, with respect to all tokens. Then, based on the values, the top $K$ tokens ($K$ being a hyper-parameter for the STRAP-ViT mechanism) with highest scores are selected and fed to the token transformation algorithm.
    \item The token transformation algorithm is designed to modify the informational content within the token sufficiently (by taking projections on a different space compared to that in which the attack would be effective), and eliminating specific features that contribute to adversarial behaviour. For example, the $L_p$-norm projection is used to curb extreme channel energy, which severely reduces adversarial potency. 
    \item Additionally, instead of choosing one transformation, we use a randomly chosen composition of transformations, to ensure that any potential future adaptive attack on STRAP-ViT becomes extremely difficult, as explained in Section \ref{discussions}. 
\end{itemize}
These transformed tokens are then sent across to the other unaffected tokens for the forward pass through the rest of the ViT model for inference. The details of this process is explained in Algorithm \ref{alg:jsd_vit_tokens_algo}, and in the subsequent subsections.


\subsection{Detection: Token Segregation}
Given the empirical observation that token entropy varies under anomaly, shown in Figure \ref{fig:entropy}, a reliable detector should measure not just “how uncertain” a token’s distribution is but “how different” it is from a clean reference. For a ViT token at layer $(\ell)$ and position $(t)$, let $(p_{\ell,t}\in\Delta^{d-1})$ be the token’s channel-wise probability vector (e.g., from $(\mathrm{softmax}(z_{\ell,t}/T)))$, and let $(q_{\ell,t})$ be a position-matched clean reference. By using the strict concavity of Shannon entropy, the \textbf{entropy gap}
\[
H\left(\tfrac12(p_{\ell,t}+q_{\ell,t})\right)-\tfrac12 H(p_{\ell,t})-\tfrac12 H(q_{\ell,t});\ge;0
\]
is zero iff $(p_{\ell,t}=q_{\ell,t})$ and positive otherwise. This quantity is exactly the Jensen–Shannon Divergence (JSD), which turns entropy’s sensitivity to anomaly into a \textit{symmetric, bounded}  $(([0,\log 2]))$ discrepancy whose square root is a metric \cite{JSD2}. Detection-theoretically, $(\mathrm{JSD}(p_{\ell,t}|q_{\ell,t})=I(U;X))$, the mutual information \cite{MI} between a latent label $(U\in{ \text{clean},\text{anomalous}})$ (equiprobable) and an observed channel $(X\sim M_{\ell,t})$; hence larger JSD means more bits of evidence that the token is anomalous \cite{JSD3}. 

Therefore, for a ViT at layer $(\ell)$ producing token embeddings $(z_{\ell,t}\in\mathbb{R}^{d})$ for tokens $(t\in{1,\dots,N})$, we have each embedding being mapped to the probability simplex via a measurable $(\phi:\mathbb{R}^{d}\to\Delta^{d-1})$; we have:
\[
p_{\ell,t}=\phi(z_{\ell,t})=\operatorname{softmax}\left(\tfrac{1}{T}z_{\ell,t}\right), 
\]
Let $(\mathcal{C})$ denote a set of clean images, and define the position- and layer-conditional reference distribution
\[
q_{\ell,t}=\mathbb{E}_{X\in\mathcal{C}}\left[p_{\ell,t}(X)\right]\in\Delta^{d-1}.
\]
With midpoint $(M_{\ell,t}=\tfrac12(p_{\ell,t}+q_{\ell,t}))$, the Jensen–Shannon Divergence (natural logs) is
\[
\mathrm{JSD}(p_{\ell,t}|q_{\ell,t})
=\tfrac12,\mathrm{KL}(p_{\ell,t}|M_{\ell,t})+\tfrac12,\mathrm{KL}(q_{\ell,t}|M_{\ell,t})
\]
\[
=H(M_{\ell,t})-\tfrac12 H(p_{\ell,t})-\tfrac12 H(q_{\ell,t})\in[0,\log 2].
\]
Anomaly strength is the metric
\[
s_{\ell,t}=\sqrt{\mathrm{JSD}(p_{\ell,t}|q_{\ell,t})}.
\]
For a target false-alarm level $(\alpha\in(0,1))$, a threshold is calibrated from clean data by
\[
\tau_{\ell}=\inf{\tau:\mathbb{P}_{X\in\mathcal{C}}(s_{\ell,t}(X)>\tau)\le \alpha},
\]
and token $(t)$ at layer $(\ell)$ is declared anomalous iff $(s_{\ell,t}>\tau_{\ell})$. 

\subsubsection{Choosing $K$ Tokens:}
While $\tau$ provides the maximum number of tokens that may be transformed for the defense technique to work, for the sake of simplicity, we introduce a hyper-parameter $K$, that is the minimum number of tokens that need to be transformed for STRAP-ViT to be effective. Different patch attacks have different sensitivities towards parts of it being modified and the patch retaining its adversarial capabilities thereafter. We have studied this over a wide range of adversarial patches of different sizes and potency and attack robustness, and concluded that it is sufficient to cover at least half ($50\%$) of the adversarial patch for the defense to work. We have observed that the requirement might be even lesser for some attack methods. This hyper-parameter $K$ is varied from $2$ tokens ($\sim$ $1\%$ of total tokens) to $4$ tokens ($\sim$ $2\%$ of total tokens) to $8$ tokens ($\sim$ $4\%$ of total tokens). It is to be noted that the transformation process does eliminate the informational content within the tokens, and so transforming many tokens, more that what is necessary, will degrade the clean case accuracy. 

\begin{algorithm}[]
\caption{STRAP-ViT Adversarial Defense with Token Segregation and Token Transformation}
\label{alg:jsd_vit_tokens_algo}
\scriptsize
\KwIn{
Image $x$; Vision Transformer with $\mathrm{PatchEmbed}$, $\mathrm{PosEnc}$, encoder blocks $\{\mathcal{B}_1,\ldots,\mathcal{B}_L\}$ and classifier head $\mathrm{Head}$; 
mapping $\phi:\mathbb{R}^d \rightarrow \Delta^{d-1}$; reference token distributions $\{q_t\}_{t=1}^N$; anomaly threshold $\tau$; 
transform parameters: $p\in\{1,2,\infty\}$, radius $r$, affine bound $\gamma<1$, softmax temperature $T$
}

\KwOut{Prediction $\hat{y}$ and transformed tokens $\{\tilde z_t\}_{t=1}^N$}

\BlankLine
\textbf{1. Tokenization} \;

$Z^{(0)} \leftarrow \mathrm{PatchEmbed}(x)$ \tcp*{$Z^{(0)}=[z^{(0)}_1,\ldots,z^{(0)}_N]$}

$Z^{(1)} \leftarrow \mathrm{PosEnc}(Z^{(0)})$ \tcp*{tokens after positional encoding}

$Z \leftarrow Z^{(1)}$ \tcp*{exclude [CLS] token}

\BlankLine
\textbf{2. Detection: Token Segregation} \;

\For{$t \leftarrow 1$ \KwTo $N$}{
    $p_t \leftarrow \phi(z_t)$ \tcp*{probability vector}
    
    $s_t \leftarrow \sqrt{\mathrm{JSD}(p_t,q_t)}$
}

$\mathcal{A} \leftarrow \{t \;|\; s_t > \tau \}$ \tcp*{set of anomalous tokens}

\BlankLine
\textbf{3. Mitigation: Randomized Token Transformation} \;

\ForEach{$t \in \mathcal{A}$}{
    
    Sample non-empty subset $S_t \subseteq \{1,2,3\}$ uniformly \;
    
    Sample permutation $\pi_t$ of $S_t$ \;
    
    $\tilde z_t \leftarrow z_t$ \;
    
    \ForEach{$k$ in order $\pi_t$}{
        
        \uIf{$k=1$}{
            $\tilde z_t \leftarrow T^{(1)}(\tilde z_t,r,p)$
        }
        \uElseIf{$k=2$}{
            $\tilde z_t \leftarrow T^{(2)}(\tilde z_t,\gamma)$
        }
        \Else{
            $\tilde z_t \leftarrow T^{(3)}(\tilde z_t,T)$
        }
    }
}

\ForEach{$t \notin \mathcal{A}$}{
    $\tilde z_t \leftarrow z_t$
}

\BlankLine
\textbf{4. ViT Forward Pass} \;

$\tilde Z^{(2)} \leftarrow \mathcal{B}_1(\tilde Z)$ \;

\For{$\ell \leftarrow 2$ \KwTo $L$}{
    $\tilde Z^{(\ell+1)} \leftarrow \mathcal{B}_\ell(\tilde Z^{(\ell)})$
}

$\hat{y} \leftarrow \mathrm{Head}(\tilde Z^{(L+1)})$

\Return{$\hat{y}, \{\tilde z_t\}_{t=1}^N$}

\end{algorithm}

\subsection{Mitigation: Token Transformation}
The objective of token transformation is to develop a scheme that is able to render the adversarial patch ineffective, by modifying the detected tokens. To ensure that the patches can not be made insensitive to such transformations at training time, we use a random selection of transformations, such that EOT operations over such transformations is not able to make the patch immune to the STRAP-ViT defense technique. After detecting anomalous tokens $(\mathcal{A}\subseteq{1,\ldots,N})$ at layer $(\ell)$, a randomized composite transform is applied token-wise. For each $(t\in\mathcal{A})$ with embedding $(z_{\ell,t}\in\mathbb{R}^d)$, a nonempty subset $(S_{\ell,t}\subseteq{1,2,3})$ and an order $(\pi_{\ell,t})$ (a permutation of $(S_{\ell,t}))$ are drawn; the transformed token is the composition
\[
\tilde z_{\ell,t}=\big(T^{(\pi_{\ell,t}(1))}\circ \cdots \circ T^{(\pi_{\ell,t}(|S_{\ell,t}|))}\big)(z_{\ell,t}),\\
\tilde z_{\ell,t}=z_{\ell,t}\ \text{for}\ t\in\mathcal{A}.
\]
The individual transformations which are available for use are mentioned here, along with how they reduce adversarial capability of the patch \cite{salomon2006transformations, yang1995projection}: 

\begin{itemize}[leftmargin=*]
    \item $L_p$-projection to curb extreme channel energy,
        \[
        T^{(1)}(z)=\Pi_{\mathbb{B}_p(r)}(z) = \arg\min_{u}{|u-z|_2:\ |u|_p\le r}
        \]
    \item Affine contraction/recentering with bounded operator norm,
        \[
        T^{(2)}(z)=A z+b,\qquad |A|_2\le \gamma<1,\ \ b\in\mathbb{R}^d,
        \] 
    \item Simplex transform via softmax to temper spikes and re-weight channels,
        \[
        T^{(3)}(z)=\operatorname{softmax}\big(z/T\big),\qquad T>0.
        \]
\end{itemize}

Parameters $((r,p,\gamma,A,b,T))$ can be randomized from designer-chosen priors to induce diversity. 

\section{Results}
\label{results}
In our evaluation, the primary focus was on assessing the model's robust accuracy under adversarial conditions. For every test setting, Top-1 accuracy refers to the percentage of test images for which the model’s single highest-probability prediction matches the true label, while Top-5 accuracy measures the percentage where the correct label appears among the model’s five most probable predictions. We noted the clean baseline accuracies, which is the performance of the model on the clean dataset. We also checked the performance of STRAP-ViT on clean samples, to study the impact of using the defense technique on clean samples and if there are many false positives. The following are the test settings:
\begin{itemize}[leftmargin=*]
    \item Models: pre-trained ViT-base-16 \cite{VIT}, DinoV2 backbone with fine tuned classifier \cite{dinov2}
    \item Datasets: ImageNet \cite{imagenet} and CalTech-101 \cite{caltech}
    \item Adversarial Attacks: 
    \begin{itemize}
        \item Single Patch: Adversarial Patch \cite{googleap}, LAVAN \cite{lavan}, GDPA \cite{gdpa} (patch sizes: 40x40, 45x45, 50x50)
        \item Multiple Patches: RP2 \cite{RP2} (two patches of size 40x40, three patches of size 35x35, four patches of size 30x30 and five patches of size 25x25)
    \end{itemize}    
\end{itemize}
All experiments were conducted on Google Colab \cite{google_colab}, with a T4 GPU support provided by Google Cloud Platform. 

Table \ref{tab1} showcases the outstanding performance of our defense technique, which significantly enhances robust accuracy against the aforementioned attacks on the aforementioned datasets and the ViT-base-16 model, with minimal impact on baseline accuracy. Notably, our defense achieves robust accuracy rates (on ImageNet dataset) of 78.8\% against Adversarial Patch and 79.2\% against LAVAN, and 77.2\% against GDPA, underscoring its effectiveness in mitigating these adversarial threats.

\newcommand{\bigval}[1]{\large{\textbf{#1}}}
\renewcommand\arraystretch{1.25}

\begin{table}[!htbp]
\caption{Performance of STRAP-ViT against adversarial patch attacks on ViT-B/16 across ImageNet and Caltech-101. Gains are shown as improvements over the no-defence accuracy.}
\label{tab1}
\resizebox{\columnwidth}{!}{
\centering
\begin{tabular}{l l l | cc | cc}
\hline
\textbf{Attack} & \textbf{Patch Size} & \textbf{Setting} 
& \textbf{ImageNet Top-1} & \textbf{Top-5} 
& \textbf{Caltech Top-1} & \textbf{Top-5} \\
\hline\hline

\multicolumn{3}{l|}{\textbf{Clean Baseline}} 
& \bigval{80.5\%} & \bigval{95.4\%} & \bigval{90.4\%} & \bigval{96.9\%} \\

\multicolumn{3}{l|}{STRAP-ViT on Clean} 
& \bigval{80.1\% (-0.4\%\,$\downarrow$)} 
& \bigval{95.1\% (-0.3\%\,$\downarrow$)} 
& \bigval{88.6\% (-1.8\%\,$\downarrow$)} 
& \bigval{95.7\% (-1.2\%\,$\downarrow$)} \\
\hline\hline

\multicolumn{7}{l}{\textbf{Adversarial Patch (GoogleAP) \cite{googleap}}} \\
\hline

& 40$\times$40 & No Defence 
& \bigval{3.6\%} & \bigval{16.4\%} & \bigval{5.3\%} & \bigval{14.9\%} \\
&        & \textbf{STRAP-ViT} 
& \bigval{78.8\% (+75.2\%\,$\uparrow$)} 
& \bigval{94.2\% (+77.8\%\,$\uparrow$)} 
& \bigval{87.6\% (+82.3\%\,$\uparrow$)} 
& \bigval{95.4\% (+80.5\%\,$\uparrow$)} \\
\hline

& 45$\times$45 & No Defence 
& \bigval{2.9\%} & \bigval{13.6\%} & \bigval{1.8\%} & \bigval{17.9\%} \\
&        & \textbf{STRAP-ViT} 
& \bigval{77.9\% (+75.0\%\,$\uparrow$)} 
& \bigval{95.3\% (+81.7\%\,$\uparrow$)} 
& \bigval{88.1\% (+86.3\%\,$\uparrow$)} 
& \bigval{94.7\% (+76.8\%\,$\uparrow$)} \\
\hline

& 50$\times$50 & No Defence 
& \bigval{1.2\%} & \bigval{9.6\%} & \bigval{1.3\%} & \bigval{5.9\%} \\
&        & \textbf{STRAP-ViT} 
& \bigval{78.3\% (+77.1\%\,$\uparrow$)} 
& \bigval{94.7\% (+85.1\%\,$\uparrow$)} 
& \bigval{87.5\% (+86.2\%\,$\uparrow$)} 
& \bigval{94.8\% (+88.9\%\,$\uparrow$)} \\
\hline\hline

\multicolumn{7}{l}{\textbf{LAVAN} \cite{lavan}} \\
\hline

& 40$\times$40 & No Defence 
& \bigval{7.4\%} & \bigval{23.9\%} & \bigval{11.2\%} & \bigval{31.1\%} \\
&        & \textbf{STRAP-ViT} 
& \bigval{79.2\% (+71.8\%\,$\uparrow$)}
& \bigval{95.3\% (+71.4\%\,$\uparrow$)}
& \bigval{88.4\% (+77.2\%\,$\uparrow$)}
& \bigval{94.6\% (+63.5\%\,$\uparrow$)} \\
\hline

& 45$\times$45 & No Defence 
& \bigval{2.1\%} & \bigval{12.8\%} & \bigval{7.9\%} & \bigval{19.2\%} \\
&        & \textbf{STRAP-ViT} 
& \bigval{78.4\% (+76.3\%\,$\uparrow$)}
& \bigval{94.3\% (+81.5\%\,$\uparrow$)}
& \bigval{89.1\% (+81.2\%\,$\uparrow$)}
& \bigval{93.8\% (+74.6\%\,$\uparrow$)} \\
\hline

& 50$\times$50 & No Defence 
& \bigval{0.3\%} & \bigval{5.4\%} & \bigval{2.4\%} & \bigval{11.3\%} \\
&        & \textbf{STRAP-ViT} 
& \bigval{78.5\% (+78.2\%\,$\uparrow$)}
& \bigval{93.9\% (+88.5\%\,$\uparrow$)}
& \bigval{87.8\% (+85.4\%\,$\uparrow$)}
& \bigval{95.4\% (+84.1\%\,$\uparrow$)} \\
\hline\hline

\multicolumn{7}{l}{\textbf{GDPA} \cite{gdpa}} \\
\hline

& 40$\times$40 & No Defence 
& \bigval{31.9\%} & \bigval{34.9\%} & \bigval{28.1\%} & \bigval{37.2\%} \\
&        & \textbf{STRAP-ViT} 
& \bigval{77.2\% (+45.3\%\,$\uparrow$)}
& \bigval{92.3\% (+57.4\%\,$\uparrow$)}
& \bigval{88.8\% (+60.7\%\,$\uparrow$)}
& \bigval{94.6\% (+57.4\%\,$\uparrow$)} \\
\hline

& 45$\times$45 & No Defence 
& \bigval{24.6\%} & \bigval{28.9\%} & \bigval{16.8\%} & \bigval{29.3\%} \\
&        & \textbf{STRAP-ViT} 
& \bigval{76.8\% (+52.2\%\,$\uparrow$)}
& \bigval{94.1\% (+65.2\%\,$\uparrow$)}
& \bigval{89.4\% (+72.6\%\,$\uparrow$)}
& \bigval{93.8\% (+64.5\%\,$\uparrow$)} \\
\hline

& 50$\times$50 & No Defence 
& \bigval{12.9\%} & \bigval{17.5\%} & \bigval{4.3\%} & \bigval{10.1\%} \\
&        & \textbf{STRAP-ViT} 
& \bigval{76.1\% (+63.2\%\,$\uparrow$)}
& \bigval{90.1\% (+72.6\%\,$\uparrow$)}
& \bigval{89.2\% (+84.9\%\,$\uparrow$)}
& \bigval{94.7\% (+84.6\%\,$\uparrow$)} \\
\hline

\end{tabular}
}
\end{table}

Table \ref{tab2} presents the performance of STRAP-ViT with the DinoV2 backbone model across datasets (ImageNet and Caltech-101), for three adversarial attacks (Adversarial Patch, LAVAN and GDPA) with varying patch sizes.

\begin{table}[!htbp]
\caption{Performance Analysis of STRAP-ViT on Image Classification Task with the DinoV2 backbone model across datasets (ImageNet and Caltech-101), for three adversarial attacks (Adversarial Patch, LAVAN and GDPA) with varying patch sizes.}
\label{tab2}
\resizebox{\columnwidth}{!}{%
\centering
\begin{tabular}{l l l | cc | cc}
\hline
\textbf{Attack} & \textbf{Patch Size} & \textbf{Setting}
& \textbf{ImageNet Top-1} & \textbf{Top-5}
& \textbf{Caltech Top-1} & \textbf{Top-5} \\
\hline\hline

\multicolumn{3}{l|}{\textbf{Clean Baseline}}
& \bigval{83.8\%} & \bigval{99.3\%} & \bigval{95.0\%} & \bigval{99.8\%} \\

\multicolumn{3}{l|}{STRAP-ViT on Clean}
& \bigval{86.0\% (+2.2\%\,$\uparrow$)}
& \bigval{97.3\% (-2.0\%\,$\downarrow$)}
& \bigval{93.5\% (-1.5\%\,$\downarrow$)}
& \bigval{96.9\% (-2.9\%\,$\downarrow$)} \\
\hline\hline

\multicolumn{7}{l}{\textbf{Adversarial Patch (GoogleAP) \cite{googleap}}} \\
\hline

& 40$\times$40 & No Defence
& \bigval{7.2\%} & \bigval{20.2\%} & \bigval{8.1\%} & \bigval{16.7\%} \\
&        & \textbf{STRAP-ViT}
& \bigval{83.9\% (+76.7\%\,$\uparrow$)}
& \bigval{96.7\% (+76.5\%\,$\uparrow$)}
& \bigval{91.3\% (+83.2\%\,$\uparrow$)}
& \bigval{98.0\% (+81.3\%\,$\uparrow$)} \\
\hline

& 45$\times$45 & No Defence
& \bigval{7.5\%} & \bigval{18.5\%} & \bigval{5.6\%} & \bigval{20.4\%} \\
&        & \textbf{STRAP-ViT}
& \bigval{83.2\% (+75.7\%\,$\uparrow$)}
& \bigval{100.3\% (+81.8\%\,$\uparrow$)}
& \bigval{91.4\% (+85.8\%\,$\uparrow$)}
& \bigval{96.4\% (+76.0\%\,$\uparrow$)} \\
\hline

& 50$\times$50 & No Defence
& \bigval{5.0\%} & \bigval{12.5\%} & \bigval{5.5\%} & \bigval{7.4\%} \\
&        & \textbf{STRAP-ViT}
& \bigval{83.8\% (+78.8\%\,$\uparrow$)}
& \bigval{97.6\% (+85.1\%\,$\uparrow$)}
& \bigval{89.6\% (+84.1\%\,$\uparrow$)}
& \bigval{96.2\% (+88.8\%\,$\uparrow$)} \\
\hline\hline

\multicolumn{7}{l}{\textbf{LAVAN} \cite{lavan}} \\
\hline

& 40$\times$40 & No Defence
& \bigval{13.0\%} & \bigval{28.1\%} & \bigval{15.4\%} & \bigval{32.2\%} \\
&        & \textbf{STRAP-ViT}
& \bigval{83.2\% (+70.2\%\,$\uparrow$)}
& \bigval{98.3\% (+70.2\%\,$\uparrow$)}
& \bigval{91.9\% (+76.5\%\,$\uparrow$)}
& \bigval{96.3\% (+64.1\%\,$\uparrow$)} \\
\hline

& 45$\times$45 & No Defence
& \bigval{6.0\%} & \bigval{17.4\%} & \bigval{11.5\%} & \bigval{22.2\%} \\
&        & \textbf{STRAP-ViT}
& \bigval{83.6\% (+77.6\%\,$\uparrow$)}
& \bigval{97.4\% (+80.0\%\,$\uparrow$)}
& \bigval{93.8\% (+82.3\%\,$\uparrow$)}
& \bigval{95.0\% (+72.8\%\,$\uparrow$)} \\
\hline

& 50$\times$50 & No Defence
& \bigval{4.1\%} & \bigval{9.5\%} & \bigval{6.3\%} & \bigval{13.3\%} \\
&        & \textbf{STRAP-ViT}
& \bigval{82.2\% (+78.1\%\,$\uparrow$)}
& \bigval{98.2\% (+88.7\%\,$\uparrow$)}
& \bigval{90.0\% (+83.7\%\,$\uparrow$)}
& \bigval{96.7\% (+83.4\%\,$\uparrow$)} \\
\hline\hline

\multicolumn{7}{l}{\textbf{GDPA} \cite{gdpa}} \\
\hline

& 40$\times$40 & No Defence
& \bigval{37.6\%} & \bigval{39.9\%} & \bigval{30.5\%} & \bigval{39.8\%} \\
&        & \textbf{STRAP-ViT}
& \bigval{81.0\% (+43.4\%\,$\uparrow$)}
& \bigval{94.3\% (+54.4\%\,$\uparrow$)}
& \bigval{91.0\% (+60.5\%\,$\uparrow$)}
& \bigval{96.8\% (+57.0\%\,$\uparrow$)} \\
\hline

& 45$\times$45 & No Defence
& \bigval{28.3\%} & \bigval{31.8\%} & \bigval{21.4\%} & \bigval{30.4\%} \\
&        & \textbf{STRAP-ViT}
& \bigval{82.2\% (+53.9\%\,$\uparrow$)}
& \bigval{97.0\% (+65.2\%\,$\uparrow$)}
& \bigval{94.0\% (+72.6\%\,$\uparrow$)}
& \bigval{94.9\% (+64.5\%\,$\uparrow$)} \\
\hline

& 50$\times$50 & No Defence
& \bigval{18.7\%} & \bigval{20.5\%} & \bigval{8.2\%} & \bigval{13.1\%} \\
&        & \textbf{STRAP-ViT}
& \bigval{80.6\% (+61.9\%\,$\uparrow$)}
& \bigval{94.8\% (+74.3\%\,$\uparrow$)}
& \bigval{93.3\% (+85.1\%\,$\uparrow$)}
& \bigval{96.7\% (+83.6\%\,$\uparrow$)} \\
\hline

\end{tabular}
}
\end{table}

We also tested STRAP-ViT on RP2 \cite{RP2}, which is a novel, more potent and stronger attack that uses multiple adversarial patches for every image. The results are reported in Table \ref{tab3}.

\begin{table}[!htbp]
\caption{Performance Analysis of STRAP-ViT on Image Classification Task with the ViT-base-16 model across datasets (ImageNet and Caltech-101), for the RP2 \cite{RP2} attack with multiple patches.}
\label{tab3}
\resizebox{\columnwidth}{!}{
\centering
\begin{tabular}{l l l | cc | cc}
\hline
\textbf{Patches} & \textbf{Patch Size} & \textbf{Setting}
& \textbf{ImageNet Top-1} & \textbf{Top-5}
& \textbf{Caltech Top-1} & \textbf{Top-5} \\
\hline\hline

\multicolumn{3}{l|}{\textbf{Clean Baseline}}
& \bigval{80.5\%} & \bigval{95.4\%} & \bigval{90.4\%} & \bigval{96.9\%} \\

\multicolumn{3}{l|}{STRAP-ViT on Clean}
& \bigval{80.1\% (-0.4\%\,$\downarrow$)}
& \bigval{95.1\% (-0.3\%\,$\downarrow$)}
& \bigval{88.6\% (-1.8\%\,$\downarrow$)}
& \bigval{95.7\% (-1.2\%\,$\downarrow$)} \\
\hline\hline

\multicolumn{7}{l}{\textbf{RP2 – TWO Patches}} \\
\hline

& 40$\times$40 & No Defence
& \bigval{31.9\%} & \bigval{34.9\%} & \bigval{28.1\%} & \bigval{37.2\%} \\
&        & \textbf{STRAP-ViT (K=4)}
& \bigval{77.2\% (+45.3\%\,$\uparrow$)}
& \bigval{92.3\% (+57.4\%\,$\uparrow$)}
& \bigval{88.8\% (+60.7\%\,$\uparrow$)}
& \bigval{94.6\% (+57.4\%\,$\uparrow$)} \\
\hline\hline

\multicolumn{7}{l}{\textbf{RP2 – THREE Patches}} \\
\hline

& 35$\times$35 & No Defence
& \bigval{24.6\%} & \bigval{28.9\%} & \bigval{16.8\%} & \bigval{29.3\%} \\
&        & \textbf{STRAP-ViT (K=4)}
& \bigval{76.8\% (+52.2\%\,$\uparrow$)}
& \bigval{94.1\% (+65.2\%\,$\uparrow$)}
& \bigval{89.4\% (+72.6\%\,$\uparrow$)}
& \bigval{93.8\% (+64.5\%\,$\uparrow$)} \\
\hline\hline

\multicolumn{7}{l}{\textbf{RP2 – FOUR Patches}} \\
\hline

& 30$\times$30 & No Defence
& \bigval{12.9\%} & \bigval{17.5\%} & \bigval{4.3\%} & \bigval{10.1\%} \\
&        & \textbf{STRAP-ViT (K=8)}
& \bigval{76.1\% (+63.2\%\,$\uparrow$)}
& \bigval{90.1\% (+72.6\%\,$\uparrow$)}
& \bigval{89.2\% (+84.9\%\,$\uparrow$)}
& \bigval{94.7\% (+84.6\%\,$\uparrow$)} \\
\hline\hline

\multicolumn{7}{l}{\textbf{RP2 – FIVE Patches}} \\
\hline

& 25$\times$25 & No Defence
& \bigval{3.1\%} & \bigval{6.3\%} & \bigval{4.8\%} & \bigval{7.9\%} \\
&        & \textbf{STRAP-ViT (K=8)}
& \bigval{73.4\% (+70.3\%\,$\uparrow$)}
& \bigval{91.2\% (+84.9\%\,$\uparrow$)}
& \bigval{86.8\% (+82.0\%\,$\uparrow$)}
& \bigval{92.1\% (+84.2\%\,$\uparrow$)} \\
\hline

\end{tabular}
}
\end{table}

Table \ref{tab4} presents the comparative study of the performance of STRAP-ViT with the state-of-the-art defense techniques. 

\begin{table}[]
\scriptsize 
\caption{Performance Comparison of STRAP-ViT with the related works, tested on the ViT model, on the ImageNet dataset with the Adversarial Patch \cite{googleap}. }
\label{tab4}
\resizebox{\columnwidth}{!}{%
\centering
\begin{tabular}{lcc|lcc}
\hline
\multicolumn{1}{c}{Defense}  & Patch-Size & Robust Acc & \multicolumn{1}{c}{Defense} & Patch-Size & Robust Acc \\ \hline \hline 
Smoothed ViT-B     \cite{salman2022certified}          & 32x32      & 39              & JEDI         \cite{jedi}                 & 40x40      & 70.6            \\
Certifiable Defense with ViT  \cite{chen2022towards}& 32x32      & 41.7            & Dimension Reduction   \cite{ChattopadhyayGH24}      & 40x40      & 62.6            \\
PatchCleanser  \cite{xiang2022patchcleanser}               & 32x32      & 62.1            & ODDR  \cite{chattopadhyay2025oddr}                       & 40x40      & 74.3            \\
LGS  \cite{naseer2019local}                        & 40x40      & 70.6            & \textbf{STRAP (ours)}                & 40x40      & \textbf{78.8}            \\ \hline
\end{tabular}
}
\end{table}

\subsection{Key Findings}       
\begin{figure}[!htbp]
\centerline{\includegraphics[width=\columnwidth]{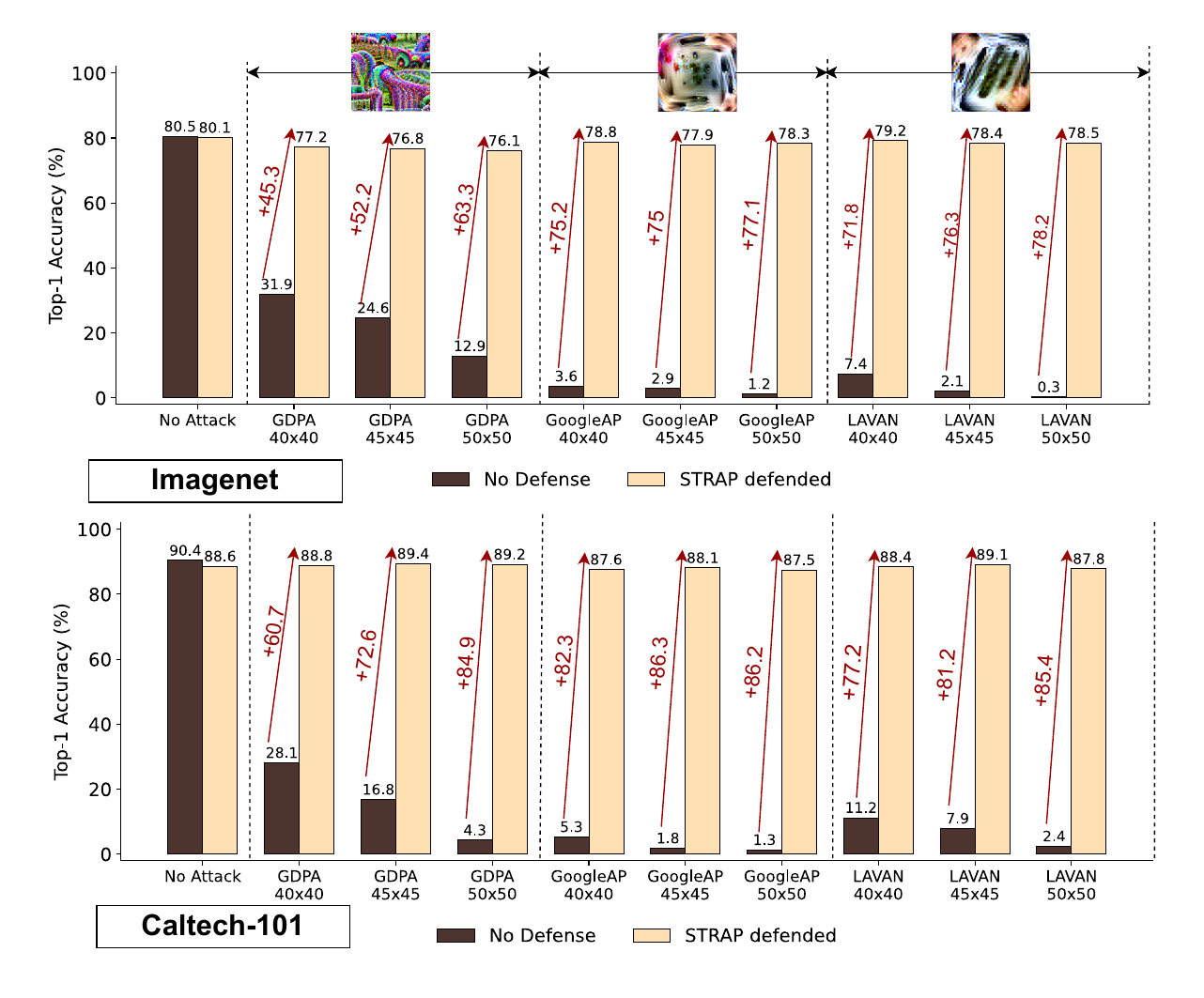}} 
\caption{A plot of the performance analysis of STRAP-ViT, with multiple adversarial patches ( Adversarial Patch, LAVAN, GDPA) and patch sizes, across two datasets (ImageNet and CalTech-101), also including the clean baseline and the impact of applying STRAP-ViT on clean samples, for reference and comparison.  }
\label{plot_1}
\end{figure}

The most important findings of the evaluation are noted here:
\begin{itemize}[leftmargin=*]
    \item As evident from Table \ref{tab1}, and shown in the Figure \ref{plot_1}, STRAP-ViT is able to provide robustness against a wide spectrum of adversarial patches (Adversarial patch, LAVAN and GDPA), of varying sizes (using the typical patch sizes as mentioned in the literature, 40x40, 45x45 and 50x50 pixel patches) and across datasets (ImageNet and CalTech-101). This establishes STRAP-ViT as a reliable method for making ViTs robust against such threats. 
    \item STRAP-ViT detects and surgically transforms only a very small portion of the tokens that are generated during the inference of the ViT models. This helps to ensure that the degradation of performance of the model, with the defense mechanism place, but in the absence of adversarial noise is minimal. This is evident from the leftmost bars in Figure \ref{plot_1}, where the performance of the model with STRAP-ViT on clean samples is very close to the clean baseline accuracy.  
    \item As a defense mechanism, STRAP-ViT is model agnostic and works for any ViT based neural architecture. This is also demonstrated in Table \ref{tab2}. 
    \item As evident from Table \ref{tab3}, STRAP-ViT is able to defend against adversarial attacks that use multiple distributed patches in one image, which is considered the strongest attack mode. It may be noted that when there are multiple patches involved, the hyper-parameter $K$ needs to be increased to account for multiple tokens overlapping with the patches present in different locations.
    \item Table \ref{tab4} substantiates the fact the STRAP-ViT outperforms other defense techniques when compared to, under similar settings. 
\end{itemize}

    


\vspace{-3mm}

\section{Discussions}
\label{discussions}
In this section, we discuss some auxiliary aspects of the STRAP-ViT defense technique, including how susceptible it is to a potential adaptive attack, tuning of hyper-parameters and the computational cost of STRAP-ViT.   

\subsection{Adaptive Attack}


Considering an attacker has full knowledge of the STRAP-ViT adversarial defense mechanism, and aims to defeat the same thereby, an adaptive attack can be implemented in the following two ways. Firstly, an adversarial patch needs to be generated that does not have informational properties through which it can be identified as an anomaly from the rest of the image. Specifically, the JSD of the adversarial patch, which is the metric of choice for STRAP-ViT, needs to be below the highest JSD of any randomly selected fragment of the clean image. We implement this by first calculating the highest JSD value among all tokens from the clean sample. Then we constrain the optimization function for the adversarial patch generation to ensure that the JSD of the patch, with respect to the rest of the clean image, does not cross the aforementioned value. Secondly, the patch needs to remain effective with the composite transforms. During the training of the patch, all possible transformations ($L_p$-transform, Affine transform and Simplex transform in this case) need to be simultaneously accounted for, so the resultant patch is immune to the defense technique \cite{eot}.

Keeping these considerations in mind, we attempted to generate the adaptive adversarial attack. We observe that such a constrained optimization problem, with the modified loss function, generates patches that have a limited adversarial attack success rate, and is able to drag down the accuracy by not more than $15\%$ as compared to the clean baseline. Additionally, the introduction of the transforms significantly reduces adversarial potency, meaning that it is very difficult to generate patches that have undergone the aforementioned transforms. 

\subsection{Hyper-parameter Tuning}
The primary hyper-parameter involved in the operations of STRAP-ViT is $K$, upon which the transformations are applied. It may be noted here that the transformations degrade the information present in the tokens, since they are targeted at eliminating noise from tokens that overlap with the adversarial patch. If $K$ is chosen to be high, then tokens which do not overlap with the adversarial patch, will also be subjected to the transformations, and this might lead to a drop in the clean case accuracy of STRAP-ViT. Therefore, we maintain the hyper-parameter $K$ at a range from $2$ tokens ($\sim$ $1\%$ of total tokens) at the lowest, to $4$ tokens ($\sim$ $2\%$ of total tokens) at an average to $8$ tokens ($\sim$ $4\%$ of total tokens) at the most.

\subsection{Computational Cost}
Let (N) be the number of tokens per image, (d) the token dimensionality, (B) the batch size, and $(A\le N)$ the number of detected anomalous tokens. The JSD-based detection maps each token $(z_t\in\mathbb{R}^d)$ to $(p_t)$ $(softmax: (O(d)))$ and evaluates $(\mathrm{JSD}(p_t|q_t))$ via the entropy form  giving $(O(Nd))$ time per image (and $(O(Nd))$ memory if probabilities are materialized; streaming reduces extra memory to $(O(d)))$. The randomized composite transformation then incurs cost only on anomalous tokens: (i) $(L_p)$ projection $(O(d))$; (ii) affine map $(Az+b)$, $(O(d))$; (iii) softmax temperature transform $(O(d))$. The STRAP-ViT defense mechanism is therefore computationally much less expensive than the transformer blocks in the ViT model itself.

\section{Conclusions}
STRAP-ViT presents a training-free modification to ViT inference that exploits statistical divergence between the information carried by an adversarial patch and the distribution of information in a clean image, enabling reliable detection and suppression of the attack. The approach is model and patch-agnostic, and provides good performance across different settings. As part of future work, this approach may be applied to Large Language Models (LLMs), particularly the Large Vision Language Models (LVLMs), which also use tokens in computer vision tasks, and can benefit from an approach like STRAP-ViT in being robust against adversarial attacks.

\section*{Acknowledgment}
This paper is partially supported by ASEAN-India Collaborative Research Project entitled "A Toolchain for Secure Hardware Accelerator Design" (CRD/2024/000829).

\bibliographystyle{ieeetr}
\bibliography{references}

@inproceedings{lavan,
  title={LaVAN: Localized and Visible Adversarial Noise},
  author={Dan Karmon}, 
  booktitle={International Conference on Machine Learning},
  year={2018}
}

@inproceedings{googleap,
title	= {Adversarial Patch},
author	= {Tom Brown},
year	= {2017},
URL	= {https://arxiv.org/pdf/1712.09665.pdf}
}

@article{levine2020randomized,
  title={(De) Randomized smoothing for certifiable defense against patch attacks},
  author={Levine, Alexander and Feizi, Soheil},
  journal={Advances in Neural Information Processing Systems},
  volume={33},
  pages={6465--6475},
  year={2020}
}

@inproceedings{xiang2021patchguard,
  title={$\{$PatchGuard$\}$: A provably robust defense against adversarial patches via small receptive fields and masking},
  author={Xiang, Chong and Bhagoji, Arjun Nitin and Sehwag, Vikash and Mittal, Prateek},
  booktitle={30th USENIX Security Symposium (USENIX Security 21)},
  pages={2237--2254},
  year={2021}
}

@inproceedings{naseer2019local,
  title={Local gradients smoothing: Defense against localized adversarial attacks},
  author={Naseer, Muzammal and Khan, Salman and Porikli, Fatih},
  booktitle={2019 IEEE Winter Conference on Applications of Computer Vision (WACV)},
  pages={1300--1307},
  year={2019},
  organization={IEEE}
}

@INPROCEEDINGS{imagenet,
  author={Deng, Jia and Dong, Wei and Socher, Richard and Li, Li-Jia and Kai Li and Li Fei-Fei},
  booktitle={2009 IEEE Conference on Computer Vision and Pattern Recognition}, 
  title={ImageNet: A large-scale hierarchical image database}, 
  year={2009},
  volume={},
  number={},
  pages={248-255},
  doi={10.1109/CVPR.2009.5206848}}

@inproceedings{cod_1,
  title={Curse of dimensionality in adversarial examples},
  author={Chattopadhyay, Nandish and Chattopadhyay, Anupam and Gupta, Sourav Sen and Kasper, Michael},
  booktitle={2019 International Joint Conference on Neural Networks (IJCNN)},
  pages={1--8},
  year={2019},
  organization={IEEE}
}

@inproceedings{cod_2,
  title={Robustness against adversarial attacks using dimensionality},
  author={Chattopadhyay, Nandish and Chatterjee, Subhrojyoti and Chattopadhyay, Anupam},
  booktitle={International Conference on Security, Privacy, and Applied Cryptography Engineering},
  pages={226--241},
  year={2021},
  organization={Springer}
}

@article{features,
  title={Spatially Correlated Patterns in Adversarial Images},
  author={Chattopadhyay, Nandish and Zhi, Lionell Yip En and Xing, Bryan Tan Bing and Chattopadhyay, Anupam},
  journal={arXiv preprint arXiv:2011.10794},
  year={2020}
}

@article{freedman,
  title={Statistics. 2007},
  author={Freedman, D and Pisani, R and Purves, R},
  journal={ISBN: 0-393970-833},
  year={1978}
}

@INPROCEEDINGS{Hu21,
  author={Hu, Yu-Chih-Tuan and Chen, Jun-Cheng and Kung, Bo-Han and Hua, Kai-Lung and Tan, Daniel Stanley},
  booktitle={2021 IEEE/CVF International Conference on Computer Vision (ICCV)}, 
  title={Naturalistic Physical Adversarial Patch for Object Detectors}, 
  year={2021},
  volume={},
  number={},
  pages={7828-7837},
  doi={10.1109/ICCV48922.2021.00775}}

@incollection{google_colab,
  title={Google colaboratory},
  author={Bisong, Ekaba},
  booktitle={Building machine learning and deep learning models on google cloud platform: a comprehensive guide for beginners},
  pages={59--64},
  year={2019},
  publisher={Springer}
}

@misc{caltech, title={Caltech 101}, DOI={10.22002/D1.20086}, abstractNote={Pictures of objects belonging to 101 categories. About 40 to 800 images per category. Most categories have about 50 images. Collected in September 2003 by Fei-Fei Li, Marco Andreetto, and Marc'Aurelio Ranzato. The size of each image is roughly 300 x 200 pixels. We have carefully clicked outlines of each object in these pictures, these are included under the 'Annotations.tar'. There is also a MATLAB script to view the annotations, 'show_annotations.m'.}, publisher={CaltechDATA}, author={Li, Fei-Fei and Andreeto, Marco and Ranzato, Marc'Aurelio and Perona, Pietro}, year={2022}, month={avr} }

@inproceedings{jedi,
  title={Jedi: Entropy-based Localization and Removal of Adversarial Patches},
  author={Tarchoun, Bilel and Ben Khalifa, Anouar and Mahjoub, Mohamed Ali and Abu-Ghazaleh, Nael and Alouani, Ihsen},
  booktitle={Proceedings of the IEEE/CVF Conference on Computer Vision and Pattern Recognition},
  pages={4087--4095},
  year={2023}
}

@inproceedings{eot,
  title={Synthesizing robust adversarial examples},
  author={Athalye, Anish and Engstrom, Logan and Ilyas, Andrew and Kwok, Kevin},
  booktitle={International conference on machine learning},
  pages={284--293},
  year={2018},
  organization={PMLR}
}

@article{Goodfellow2015ExplainingAH,
  title={Explaining and Harnessing Adversarial Examples},
  author={I. J. Goodfellow and J. Shlens and C. Szegedy},
  journal={CoRR},
  year={2015},
  volume={abs/1412.6572}}

@inproceedings{VIT,
  title={Tokens-to-token vit: Training vision transformers from scratch on imagenet},
  author={Yuan, Li and Chen, Yunpeng and Wang, Tao and Yu, Weihao and Shi, Yujun and Jiang, Zi-Hang and Tay, Francis EH and Feng, Jiashi and Yan, Shuicheng},
  booktitle={Proceedings of the IEEE/CVF international conference on computer vision},
  pages={558--567},
  year={2021}
}

@inproceedings{swin,
  title={Swin transformer: Hierarchical vision transformer using shifted windows},
  author={Liu, Ze and Lin, Yutong and Cao, Yue and Hu, Han and Wei, Yixuan and Zhang, Zheng and Lin, Stephen and Guo, Baining},
  booktitle={Proceedings of the IEEE/CVF international conference on computer vision},
  pages={10012--10022},
  year={2021}
}

@inproceedings{ChattopadhyayGH24,
  author       = {Nandish Chattopadhyay and
                  Amira Guesmi and
                  Muhammad Abdullah Hanif and
                  Bassem Ouni and
                  Muhammad Shafique},
  title        = {Defending against Adversarial Patches using Dimensionality Reduction},
  booktitle    = {{DAC}},
  pages        = {222:1--222:6},
  publisher    = {{ACM}},
  year         = {2024}
}

@inproceedings{chattopadhyay2025oddr,
  title={Oddr: Outlier detection \& dimension reduction based defense against adversarial patches},
  author={Chattopadhyay, Nandish and Guesmi, Amira and Hanif, Muhammad Abdullah and Ouni, Bassem and Shafique, Muhammad},
  booktitle={Proceedings of the IEEE/CVF International Conference on Computer Vision},
  pages={22999--23008},
  year={2025}
}

@article{MI,
  title={Estimating mutual information},
  author={Kraskov, Alexander and St{\"o}gbauer, Harald and Grassberger, Peter},
  journal={Physical Review E—Statistical, Nonlinear, and Soft Matter Physics},
  volume={69},
  number={6},
  pages={066138},
  year={2004},
  publisher={APS}
}

@inproceedings{cod3,
  title={Robust Perception for Autonomous Vehicles using Dimensionality Reduction},
  author={Garg, Shivam and Chattopadhyay, Nandish and Chattopadhyay, Anupam},
  booktitle={2022 IEEE International Conference on Trust, Security and Privacy in Computing and Communications (TrustCom)},
  pages={1516--1521},
  year={2022},
  organization={IEEE}
}

@article{chattopadhyay2024anomaly,
  title={Anomaly Unveiled: Securing Image Classification against Adversarial Patch Attacks},
  author={Chattopadhyay, Nandish and Guesmi, Amira and Shafique, Muhammad},
  journal={arXiv preprint arXiv:2402.06249},
  year={2024}
}

@article{chiang2020certified,
  title={Certified defenses for adversarial patches},
  author={Chiang, Ping-yeh and Ni, Renkun and Abdelkader, Ahmed and Zhu, Chen and Studer, Christoph and Goldstein, Tom},
  journal={arXiv preprint arXiv:2003.06693},
  year={2020}
}

@inproceedings{chen2022towards,
  title={Towards practical certifiable patch defense with vision transformer},
  author={Chen, Zhaoyu and Li, Bo and Xu, Jianghe and Wu, Shuang and Ding, Shouhong and Zhang, Wenqiang},
  booktitle={Proceedings of the IEEE/CVF Conference on Computer Vision and Pattern Recognition},
  pages={15148--15158},
  year={2022}
}

@inproceedings{xiang2022patchcleanser,
  title={$\{$PatchCleanser$\}$: Certifiably robust defense against adversarial patches for any image classifier},
  author={Xiang, Chong and Mahloujifar, Saeed and Mittal, Prateek},
  booktitle={31st USENIX security symposium (USENIX Security 22)},
  pages={2065--2082},
  year={2022}
}

@inproceedings{gittings2020vax,
  title={Vax-a-net: Training-time defence against adversarial patch attacks},
  author={Gittings, Thomas and Schneider, Steve and Collomosse, John},
  booktitle={Proceedings of the Asian Conference on Computer Vision},
  year={2020}
}

@inproceedings{rao2020adversarial,
  title={Adversarial training against location-optimized adversarial patches},
  author={Rao, Sukrut and Stutz, David and Schiele, Bernt},
  booktitle={European conference on computer vision},
  pages={429--448},
  year={2020},
  organization={Springer}
}

@inproceedings{hayes2018visible,
  title={On visible adversarial perturbations \& digital watermarking},
  author={Hayes, Jamie},
  booktitle={Proceedings of the IEEE conference on computer vision and pattern recognition workshops},
  pages={1597--1604},
  year={2018}
}

@inproceedings{salman2022certified,
  title={Certified patch robustness via smoothed vision transformers},
  author={Salman, Hadi and Jain, Saachi and Wong, Eric and Madry, Aleksander},
  booktitle={Proceedings of the IEEE/CVF conference on computer vision and pattern recognition},
  pages={15137--15147},
  year={2022}
}

@article{gdpa,
  title={Generative dynamic patch attack},
  author={Li, Xiang and Ji, Shihao},
  journal={arXiv preprint arXiv:2111.04266},
  year={2021}
}

@article{dinov2,
  title={Dinov2: Learning robust visual features without supervision},
  author={Oquab, Maxime and Darcet, Timoth{\'e}e and Moutakanni, Th{\'e}o and Vo, Huy and Szafraniec, Marc and Khalidov, Vasil and Fernandez, Pierre and Haziza, Daniel and Massa, Francisco and El-Nouby, Alaaeldin and others},
  journal={arXiv preprint arXiv:2304.07193},
  year={2023}
}

@misc{BI,
  author = {Business Insider},
  title = {Big Tech's AI spending spree just keeps getting bigger},
  howpublished = {\url{https://www.businessinsider.com/big-tech-capex-spending-ai-earnings-2025-10}},
  year = {2025},
  note = {Accessed: November 17, 2025} 
}

@misc{investopedia,
  author = {Investopedia},
  title = {Amazon Follows Google, Meta, and Microsoft With Plans To Boost Spending on AI},
  howpublished = {https://www.investopedia.com/amazon-follows-google-meta-and-microsoft-with-plans-to-boost-spending-on-ai-8787507},
  year = {2025},
  note = {Accessed: November 17, 2025} 
}

@book{entropy,
  title={Entropy and information theory},
  author={Gray, Robert M},
  year={2011},
  publisher={Springer Science \& Business Media}
}

@article{JSD1,
  title={The jensen-shannon divergence},
  author={Men{\'e}ndez, Mar{\'\i}a Luisa and Pardo, Julio Angel and Pardo, Leandro and Pardo, Mar{\'\i}a del C},
  journal={Journal of the Franklin Institute},
  volume={334},
  number={2},
  pages={307--318},
  year={1997},
  publisher={Elsevier}
}

@inproceedings{JSD2,
  title={Jensen-Shannon divergence and Hilbert space embedding},
  author={Fuglede, Bent and Topsoe, Flemming},
  booktitle={International symposium onInformation theory, 2004. ISIT 2004. Proceedings.},
  pages={31},
  year={2004},
  organization={IEEE}
}

@article{JSD3,
  title={On a generalization of the Jensen--Shannon divergence and the Jensen--Shannon centroid},
  author={Nielsen, Frank},
  journal={Entropy},
  volume={22},
  number={2},
  pages={221},
  year={2020},
  publisher={MDPI}
}

@book{salomon2006transformations,
  title={Transformations and projections in computer graphics},
  author={Salomon, David},
  year={2006},
  publisher={Springer}
}

@article{yang1995projection,
  title={Projection-based spatially adaptive reconstruction of block-transform compressed images},
  author={Yang, Yongyi and Galatsanos, Nikolas P and Katsaggelos, Aggelos K},
  journal={IEEE transactions on image processing},
  volume={4},
  number={7},
  pages={896--908},
  year={1995},
  publisher={IEEE}
}

@inproceedings{RP2,
  title={Robust physical-world attacks on deep learning visual classification},
  author={Eykholt, Kevin and Evtimov, Ivan and Fernandes, Earlence and Li, Bo and Rahmati, Amir and Xiao, Chaowei and Prakash, Atul and Kohno, Tadayoshi and Song, Dawn},
  booktitle={Proceedings of the IEEE conference on computer vision and pattern recognition},
  pages={1625--1634},
  year={2018}
}


\end{document}